\documentclass[conference]{IEEEtran}

\usepackage{graphicx}
\usepackage{caption}
\usepackage{subcaption}
\usepackage{float}
\usepackage[bottom]{footmisc}

\usepackage{multicol}
\usepackage{longtable}
\usepackage[table]{xcolor}
\usepackage{cite}
\usepackage{algorithm}
\usepackage{algorithmic}
\usepackage[inline]{enumitem}
\usepackage{amsmath}

\begin{document}

\author{
Somnath Rakshit \\
Jalpaiguri Govt. Engg. College\\
\texttt{somnath52@gmail.com}\\
\and
Soumyadeep Debnath\\
Jalpaiguri Govt. Engg. College\\
\texttt{soumyadebnath13@gmail.com}\\
\and
Dhiman Mondal\\
Jalpaiguri Govt. Engg. College\\
\texttt{mondal.dhiman@gmail.com}\\
}
\title{Identifying Land Patterns from Satellite Imagery in Amazon Rainforest using Deep Learning }

\maketitle

\begin{abstract}
\boldmath
The Amazon rainforests have been suffering widespread damage, both via natural and artificial means. Every minute, it is estimated that the world loses forest cover the size of 48 football fields. Deforestation in the Amazon rainforest has led to drastically reduced biodiversity, loss of habitat, climate change and other biological losses. In this respect, it has become essential to track how the nature of these forests change over time. Image classification using deep learning can help speed up this process by removing the manual task of classifying each image. Here, it is shown how convolutional neural networks can be used to track changes in land patterns in the Amazon rainforests. In this work, a testing accuracy of 96.71\% was obtained. This can help governments and other agencies to track changes in land patterns more effectively and accurately.

\end{abstract}



\section{Introduction}

\IEEEPARstart{T}{he} Amazon rainforest is a moist broadleaf forest in South America that covers a land area of more than 7,000,000 km\textsuperscript{2}. There are nine countries that fall under the Amazon rainforests including countries like Brazil, Peru, Columbia, Venezuela, Ecuador, Bolivia etc. as mentioned in  \textit{Fig. 1}. More than half of the world’s total rainforest cover is represented by the Amazon. World’s largest and most biologically diverse tract of tropical rainforest is thought to be present here with almost 400 billion individual trees of more than 15,000 species. The Amazon rainforest is thought to be in existence for at least 55 million years and is richer in wildlife biodiversity too than the African or Asian wet forests.\\

\begin{figure}[H]
\centering
\includegraphics[width=50mm]{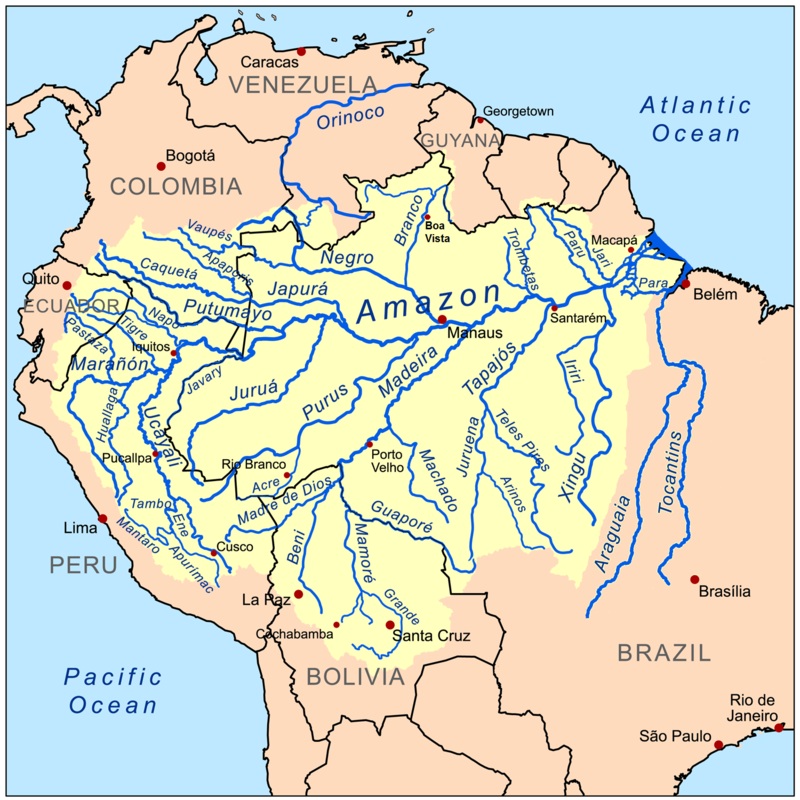}
\caption{Map of the Amazon Basin }
\label{img_amazon}
\end{figure}

The Brazilian part of the Amazon was largely intact till the construction of the Trans-Amazonian Highway began in the 1970s ~\cite{Nunes2012}. It triggered a high level of deforestation and since then, this rate has fluctuated but has always remained high consistently as shown in  \textit{Fig. 2} 
\footnote[1]{https://rainforests.mongabay.com/amazon/amazon\_destruction.html}
. This makes it important to develop an automated approach of monitoring the land pattern in the Amazon rainforests that will help the governmental agencies to easily and effectively monitor the deforestation and other activities taking place in the Amazon.\\

\begin{figure}[H]
\centering
\includegraphics[width=80mm]{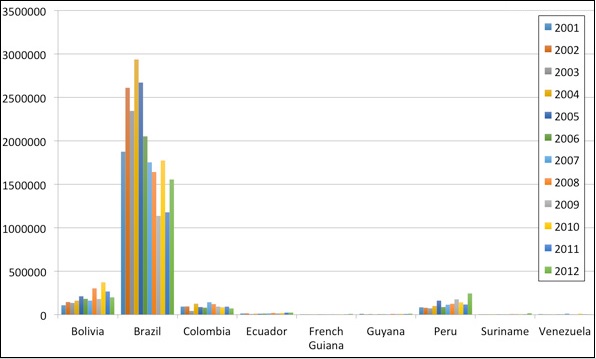}
\caption{Annual Forest Loss in the Amazon}
\label{img_loss}
\end{figure}

Computer vision is an important field these days and deep learning using neural networks have been at the forefront of it. Using a wide range of deep learning techniques, it has been possible to achieve higher accuracy than the previous models by a noticeable amount. Also, deep learning triumphs conventional methods of computer vision by not requiring hand-crafted features. Thus, this is a technique that can be applied to a wide variety of problems. In the proposed work, an automated way to label satellite images with their corresponding class of land cover has been developed. The VGG16 model has been used after fine tuning and achieved high accuracy. This work may be extended by using other well-known image classification models. Deforestation in the Amazon rainforests has increased considerably in the past few decades. This has affected its biodiversity and climate adversely. Using satellite imagery, it has been possible to track the changes taking place within a region. However, due to the presence of huge amount of data, it requires considerable manual resources for proper labelling.

\section{RELATED WORKS}
Neural networks have been used to solve a wide range of problems. It has been quite successfully used in problems such as the prediction of diabetes ~\cite{Somnath2017}, face detection ~\cite{Rowley1998}, object localization ~\cite{Pierre2013}, etc. Visual image recognition has gained spectacular popularity in the current scenario owing to the tremendously fast-paced research on this topic. Deep convolutional networks started gaining popularity from the year 2012 when Krizhevsky et al.'s ~\cite{Krizhevsky2012} AlexNet won the ImageNet challenge and beat other models by a big margin. Since then, all winners of ImageNet challenge have used a deep convolutional neural network architecture. Simonyan and Zisserman ~\cite{Simonyan2014} developed the VGG model that improved the error rate in the ImageNet challenge.\\
However, deep learning was not mathematically understood well by the community. In this regard, the works of Zeiler at al. ~\cite{Zeiler2014} helped in throwing light in how deep learning models work. Here, it was shown that how deep learning models learned features with respect to each layer.\\
 Rajat et al. ~\cite{Raina2007} introduced the concept of transfer learning in the year 2007. Using this, it is possible to use the weights of the model that was trained on a particular dataset and retrain the last layer before passing its output through a classifier. This allows to get the results that are nearly as good as the successful models without requiring the enormous computing resources that were used in training the original model.\\
Gardner et al. ~\cite{Gardner}  used the ResNet50 model to classify land patterns in the Amazon rainforests and obtained a F-Score of 0.91. They used a number of data augmentation and ensemble techniques for this purpose.\\
Longwell et al. ~\cite{Longwell} used the near-IR channel of the image and through a deep residual architecture, obtained a F2 score of 0.9. \\
To speed up the development using neural networks, quite a lot software libraries have been developed. Some examples include Scikit-Learn \cite{Pedregosa2011}, Tensorflow \cite{Abadi2016}, Keras ~\cite{Chollet2015} for machine intelligence and Pandas \cite{McKinney2010, McKinney1996} for data analysis. These libraries are often used not only in deep learning but also in many other tasks.

\section{DATASET}
The dataset for this work has been derived from Planet’s full-frame analytic scene products using its 4-band satellites in sun-synchronous orbit (SSO) and International Space Station (ISS) orbit.
\subsection{Chip (Image) Data Format}

The set of chips were captured with four bands of data each, viz. red, green, blue and near-infrared. The GeoTIFF data was originally captured along with the chips. However, they have been removed with ground control points (GCPs) for the purpose of  this experiment as the data was not found to be essential.\\
The images in this dataset have a ground-sample distance (GSD) of 3.7 m and orthorectified pixel size of 3 m. Planet’s Flock 
\footnote[2]{https://www.planet.com/docs/spec-sheets/sat-imagery/\#ps-imagery-product}
satellites have been used to collect this data in both sun-synchronous and ISS orbits between January 1, 2016 and February 1, 2017. In the dataset, the TIFF files were converted to JPG files for easier processing using the Planet visual product processor .

\subsection{Data Labeling Process and Quality}
The Crowd Flower platform 
\footnote[3]{https://www.figure-eight.com/} was used to label this dataset using crowdsourced labour. Planet acknowledges the fact that although the utmost care was taken to correctly label the dataset, not all labels are accurate in nature.\\
While labelling, the whole dataset was divided into two sets, a "hard" and an "easy" set. Scenes having easier-to-identify labels like primary rainforest, agriculture, habitation, roads, water, and cloud conditions were placed in the easy category. Shifting cultivation, slash and burn agriculture, blow down, mining, and other phenomena were placed in the hard category.

\subsection{Class Labels}
Planet's Impact team
\footnote[4]{https://www.planet.com/markets/impact/}
was consulted while labelling the dataset. This dataset reasonably represents the places of interest in the Amazon basin. The labels can broadly be broken into three groups. They are:
\begin{figure}[H]
	\centering
	\begin{subfigure}[b]{0.15\textwidth}
		\includegraphics[width=25mm]{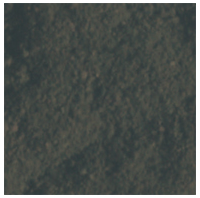}
		\caption{}
		\label{img1}
	\end{subfigure}
	\hfill
	\begin{subfigure}[b]{0.15\textwidth}
		\includegraphics[width=25mm]{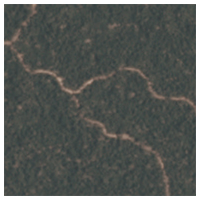}
		\caption{}
		\label{img2}
	\end{subfigure}
	\hfill
	\begin{subfigure}[b]{0.15\textwidth}
		\includegraphics[width=25mm]{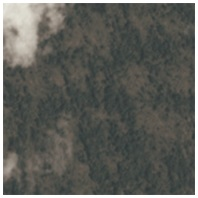}
		\caption{}
		\label{img3}
	\end{subfigure}
	
	\begin{subfigure}[b]{0.15\textwidth}
		\includegraphics[width=25mm]{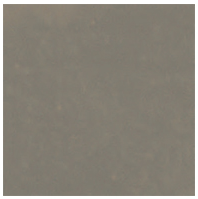}
		\caption{}
		\label{img4}
	\end{subfigure}
	\hfill
	\begin{subfigure}[b]{0.15\textwidth}
		\includegraphics[width=25mm]{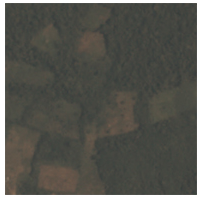}
		\caption{}
		\label{img5}
	\end{subfigure}
	\hfill
	\begin{subfigure}[b]{0.15\textwidth}
		\includegraphics[width=25mm]{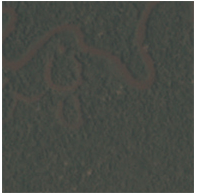}
		\caption{}
		\label{img6}
	\end{subfigure}
	
	\begin{subfigure}[b]{0.15\textwidth}
		\includegraphics[width=25mm]{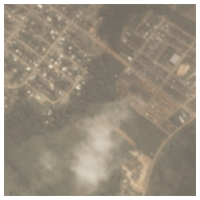}
		\caption{}
		\label{img7}
	\end{subfigure}
	\hfill
	\begin{subfigure}[b]{0.15\textwidth}
		\includegraphics[width=25mm]{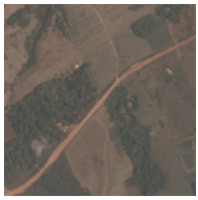}
		\caption{}
		\label{img8}
	\end{subfigure}
	\hfill
	\begin{subfigure}[b]{0.15\textwidth}
		\includegraphics[width=25mm]{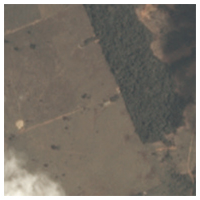}
		\caption{}
		\label{img9}
	\end{subfigure}
	\caption{Sample Chips and Their Labels (a) primary (b) roads + primary (c) partly\_cloudy + primary (d) haze + primary (e) cultivation + primary (f) water + primary (g) habitation + partly\_cloudy (h) agriculture + roads + primary (i) agriculture + pasture + primary + partly\_cloudy}
	\label{img_grid}
\end{figure}
\begin{enumerate}
\item Atmospheric Conditions
\item Common Land Cover (Land Use Phenomena)
\item Rare Land Cover (Land Use Phenomena)
\end{enumerate}

The whole dataset has 17 type of labels which had to be identified for each of the chips showed in Figure \ref{img_grid}. They are described below:

\begin{multicols}{2}
\begin{itemize}
\item agriculture   
\item artisinal\_mine
\item bare\_ground			
\item blooming
\item blow\_down			
\item clear
\item cloudy			
\item cultivation
\item habitation			
\item haze
\item partly\_cloudy		
\item primary
\item road				
\item selective\_logging
\item conventional\_mine		
\item slash\_burn
\item water
\end{itemize}
\end{multicols}

As a single image can have multiple classes in this dataset so, in the algorithm, all such classes were tried to predict correctly for each of the images.

\section{DATA ANALYSIS}
Some basic data analysis was performed on the dataset which have been described in details below.

\subsection{Distribution of Training Labels}
\begin{figure}[H]
\centering
\includegraphics[width=90mm]{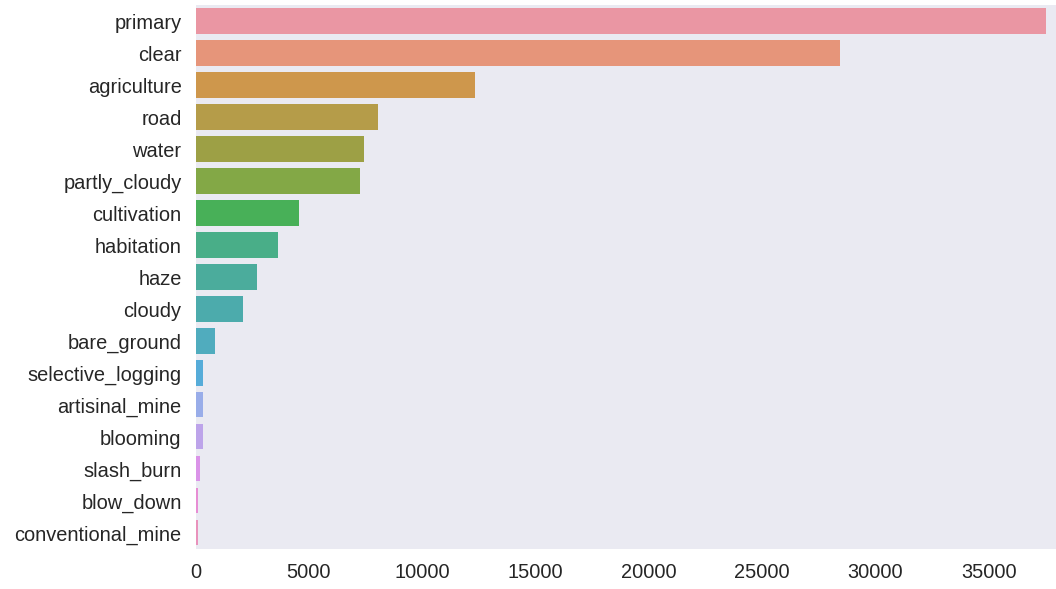}
\caption{Distribution of Training Labels}
\label{img_train}
\end{figure}

Firstly, the histogram as present in Figure \ref{img_train} showing the distribution of training labels was constructed. It has been found that the dataset is not balanced in nature, i.e.\ , all labels are not present in uniform quantity. Labels such as primary, clear and agriculture are present in significantly more number than the other ones. Whereas, some other labels like slash\_burn, blow\_down and conventional\_mine are present in very less quantity. Note that in the dataset, a single image may have multiple classes. The histogram must be seen keeping this in mind.

\subsection{Correlation Matrix}
\begin{figure}[H]
\centering
\includegraphics[width=90mm]{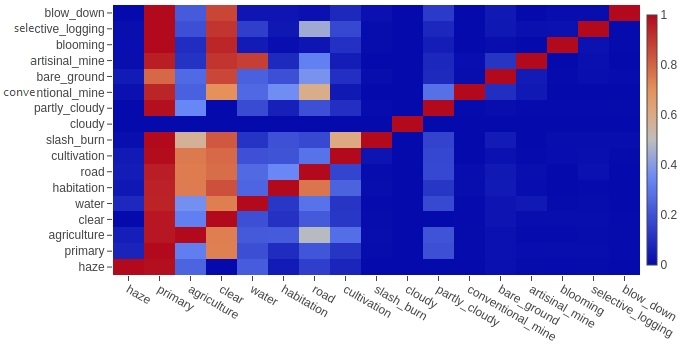}
\caption{Distribution of Training Labels}
\label{img_dist}
\end{figure}

The correlation matrix was plotted, as shown in Figure \ref{img_dist}, to understand the occurrence of the classes with respect to each other. Here, redder is the label, more is the value of the correlation for any given pair of classes. After studying this plot, some interesting results were observed. Some of them are:
\begin{itemize}
\item The label primary is associated with almost all classes. This means that most chips have some degree of primary forests along with other labels.
\item The label agriculture is also associated with a few labels like road, habitation and cultivation.
\end{itemize} 

\section{PREPROCESSING OF DATASET}
Even after converting to JPG, the dataset was quite large in size. It would have been computationally expensive to train the model on such a large dataset. Besides, the obtained dataset contained images of various dimensions. Hence, all images were resized to a standard size, in this case, 128x128 pixels. This is also an important step as it helps in speeding up the training.  Since the downloaded VGG16 model did not contain the top layer, it was possible to train with images with dimensions (128x128x3) that were different from the dimensions of images used in the original VGG16 model (224x224x3). In this dataset, 40479 images for training and 40669 images for testing were used. Each image may be classified into multiple classes.

\section{METHODOLOGY}
\begin{figure}[H]
\centering
\includegraphics[width=90mm]{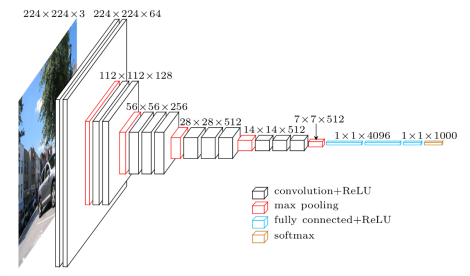}
\caption{Distribution of Training Labels}
\label{img_vgg}
\end{figure}

In the proposed work, the VGG16 model has been used to classify images into various classes. Figure \ref{img_vgg} shows the original diagram of the VGG model.

In the model, a batch normalization layer was added to the input layer and then fed to the VGG16 model. The last block of the original VGG16 model was removed and the output of the penultimate block of the VGG16 model was flattened. It was then passed on to a softmax classifier to present the output with respect to 17 classes. Here, 20\% of the training data was used for validation after training. The architecture of this model is present in Table \ref{tab_model}. \\

\begin{table}[H]
\caption{Architecture of VGG16 model}
\begin{center}
\begin{tabular}{|p{2.5cm}|p{2.5cm}|p{1.5cm}|}
\hline
Layer (type) & Output Shape & Parameter\\
\hline
input\_1 (InputLayer) & (None, 128, 128, 3) & 0 \\
\hline
batch\_normalization\_1 & (None, 128, 128, 3) & 12 \\
\hline
vgg16 (Model)  & (None, 4, 4, 512) & 14714688 \\
\hline
flatten\_1 (Flatten) & (None, 8192) & 0 \\
\hline
dense\_1 (Dense) & (None, 17) & 139281 \\
\hline
\end{tabular}
\end{center}
\label{tab_model}
\end{table}

\begin{figure}[H]
\centering
\includegraphics[width=90mm]{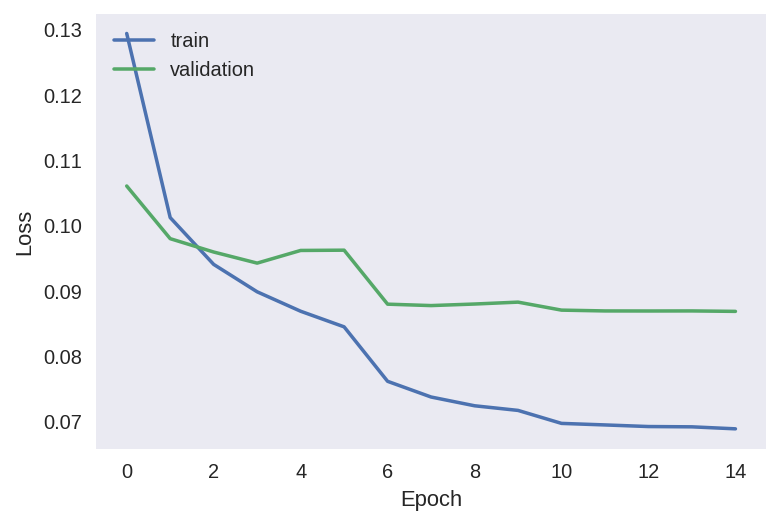}
\caption{Plot of Training Loss vs Epoch}
\label{img_plot}
\end{figure}

Here, the Adam optimizer ~\cite{Kingma2014} has been used to minimize the loss, which is measured by binary cross- entropy, with a learning rate of 10−4. Batch size of 128 was used here and this model was trained for 15 epochs. By this time, the training loss had converged. Using an NVIDIA Tesla K80 GPU, this took around one hour to train. The plot between the training loss vs epoch is shown in Figure \ref{img_plot}.

\section{RESULT}

The following metrics were evaluated in our work

\begin{equation*}
     \text{Precision} = \frac{TP}{TP+FP}
\end{equation*}

\begin{equation*}
    \text{Recall} = \frac{TP}{TP+FN}
\end{equation*}

\begin{equation*}
    \text{Accuracy} = \frac{TP+TN}{TP+TN+FP+FN}
\end{equation*}

With \textit{TP}, \textit{FP}, \textit{TN},\textit{FN} being number of true positives, false positives, true negatives and false negatives, respectively.

\begin{equation*}
\begin{split}
\text{F-Beta Score} = F_{\beta}= \frac{1}{\frac{1}{\beta +1}\frac{1}{precision}+\frac{\beta}{\beta +1}\frac{1}{recall}}\\
=(1+\beta)\frac{precision.recall}{precision+recall}
\end{split}
\end{equation*}
\begin{equation*}
\text{Categorical Cross Entropy} = \sum_{i}^{n}\sum_{k}^{K} {-y}_{true}^{(k)} log(y_{predict}^{(k)})
\end{equation*}

In the experiment, a training loss of 6.88\%, training accuracy of 97.35\% and testing accuracy of 96.71\% were obtained.\\
Also, an F-beta score of 92.69\% was obtained. The F-beta score is a weighted harmonic mean of the precision and recall. An F-beta score reaches its best value at 1 and worst score at 0.
 \section{CONCLUSION}
 In this work, a way to classify satellite imagery in an automated manner using deep learning with the help of the VGG16 model has been shown. High accuracy was consuming one hour while training with an NVIDIA Tesla K80 GPU. This model can be successfully applied to track the changing land pattern in the rainforests of Amazon. This data about the location of deforestation and human encroachment on forests can help governments and local stakeholders respond more quickly and effectively. Besides, this model can be used to track natural calamities like floods, forest fires, etc.

\section{FUTURE SCOPE}

A few additions may be made to this work for improvements mentioned below:\\ \\
Using a larger neural network is likely to give a better result. Models like ResNet and Inception, which are deeper in nature may give better results than the VGG16 model.\\ \\
Also, increased preprocessing of the dataset may help in better classification. In this work, it has been shown how resizing the provided image to 128x128 pixels can be made to obtain good performance. No preprocessing involving the texture and nature of the image itself was performed.\\ \\
Performing data augmentation to make the system more robust may be another way of getting better results. Since the satellite images may vary in terms of lighting effect, rotation, shifting, etc., it may be a good idea to perform data augmentation to enlarge the dataset for better training.\\ \\
These things may be investigated in the upcoming future to improve the accuracy and robustness of this model.

\bibliography{coseds}

\begin{thebibliography}{10}
\providecommand{\url}[1]{#1}
\csname url@samestyle\endcsname
\providecommand{\newblock}{\relax}
\providecommand{\bibinfo}[2]{#2}
\providecommand{\BIBentrySTDinterwordspacing}{\spaceskip=0pt\relax}
\providecommand{\BIBentryALTinterwordstretchfactor}{4}
\providecommand{\BIBentryALTinterwordspacing}{\spaceskip=\fontdimen2\font plus
\BIBentryALTinterwordstretchfactor\fontdimen3\font minus
  \fontdimen4\font\relax}
\providecommand{\BIBforeignlanguage}[2]{{%
\expandafter\ifx\csname l@#1\endcsname\relax
\typeout{** WARNING: IEEEtran.bst: No hyphenation pattern has been}%
\typeout{** loaded for the language `#1'. Using the pattern for}%
\typeout{** the default language instead.}%
\else
\language=\csname l@#1\endcsname
\fi
#2}}
\providecommand{\BIBdecl}{\relax}
\BIBdecl

\bibitem{Nunes2012}
{Nunes Kehl, Thiago, Viviane Todt, Mauricio Roberto Veronez, and Silvio César
  Cazella}, ``{Amazon rainforest deforestation daily detection tool using
  artificial neural networks and satellite images},'' \emph{{Sustainability
  4}}, vol.~10, pp. 2566--2573, 2012.

\bibitem{Somnath2017}
{Somnath Rakshit, Suvojit Manna, Sanket Biswas, RiyankaKundu, Priti Gupta,
  Sayantan Maitra, and Subhas Barman}, ``{Prediction of Diabetes Type-II Using
  a Two-Class Neural Network},'' \emph{{International Conference on
  Computational Intelligence, Communications, and Business
  Analytics,Springer,Singapore}}, pp. 65--71, 2017.

\bibitem{Rowley1998}
{Rowley, Henry A., Shumeet Baluja, and Takeo Kanade}, ``{Neural network-based
  face detection},'' \emph{{IEEE Transactions on pattern analysis and machine
  intelligence}}, vol.~20, no.~1, pp. 23--28, 1998.

\bibitem{Pierre2013}
{ Pierre Sermanet, David Eigen, Xiang Zhang, Michaël Mathieu, Rob Fergus, and
  Yann LeCun}, ``{Overfeat: Integrated recognition, localization and detection
  using convolutional networks.}'' \emph{{arXiv preprint arXiv:1312.6229 }},
  2013.

\bibitem{Krizhevsky2012}
{Krizhevsky, Alex, Ilya Sutskever, and Geoffrey E. Hinton.}, ``{Imagenet
  classification with deep convolutional neural networks},'' \emph{{Advances in
  neural information processing systems}}, pp. 1097--1105, 2012.

\bibitem{Simonyan2014}
{Simonyan, Karen, and Andrew Zisserman}, ``{Very deep convolutional networks
  for large-scale image recognition},'' \emph{{arXiv preprint arXiv:1409.1556
  }}, 2014.

\bibitem{Zeiler2014}
{Zeiler, Matthew D., and Rob Fergus}, ``{Visualizing and understanding
  convolutional networks},'' \emph{{European conference on computer
  vision,Springer, Cham}}, pp. 818--833, 2014.

\bibitem{Raina2007}
{Raina, Rajat, Alexis Battle, Honglak Lee, Benjamin Packer, and Andrew Y. Ng.
  }, ``{Self-taught learning: transfer learning from unlabeled data.}''
  \emph{{Proceedings of the 24th international conference on Machine
  learning"}, volume =}.

\bibitem{Gardner}
{Gardner, Daniel, and David Nichols}, ``{Multi-label Classification of
  Satellite Images with Deep Learning}.''

\bibitem{Longwell}
{Longwell, Scott, Tyler Shimko and Alex Williams}, ``{DeepRootz: Classifying
  satellite images of the Amazon rainforest},'' \emph{{}}.

\bibitem{Pedregosa2011}
{Pedregosa, Fabian, Gaël Varoquaux, Alexandre Gramfort, Vincent Michel,
  Bertrand Thirion, Olivier Grisel, Mathieu Blondel et al.}, ``{Scikit-learn:
  Machine learning in Python.}'' \emph{{Journal of machine learning research}},
  vol.~12, pp. 2825--2830, 2011.

\bibitem{Abadi2016}
{Abadi, Martín, Paul Barham, Jianmin Chen, Zhifeng Chen, Andy Davis, Jeffrey
  Dean, Matthieu Devin et al. }, ``{TensorFlow: A System for Large-Scale
  Machine Learning.}'' \emph{{In OSDI}}, vol.~16, pp. . 265--283, 2016.

\bibitem{Chollet2015}
{Chollet, François}, ``{Keras},'' \emph{{}}, 2015.

\bibitem{McKinney2010}
{McKinney, Wes. }, ``{Data structures for statistical computing in python.}''
  \emph{{Proceedings of the 9th Python in Science Conference}}, vol. 445, pp.
  51--56, 2010.

\bibitem{McKinney1996}
{McKinney, Wes}, ``{pandas: a foundational Python library for data analysis and
  statistics},'' \emph{{Python for High Performance and Scientific Computing
  }}, pp. 1--9, 2011.

\bibitem{Kingma2014}
{Kingma, Diederik P., and Jimmy Ba}, ``{Adam: A method for stochastic
  optimization},'' \emph{{arXiv preprint arXiv:1412.6980}}, 2014.

\end{thebibliography}

\end{document}